%% file: samplepaper.tex
\documentclass[runningheads]{llncs}
\usepackage{graphicx}
\usepackage{tikz}
\usepackage{comment}
\usepackage{amsmath,amssymb} 
\usepackage{color}
\usepackage{wrapfig}
\usepackage{booktabs}
\usepackage[accsupp]{axessibility}  

\usepackage{algorithm}
\usepackage{algpseudocode}

\AtBeginDocument{
  \setlength\abovedisplayskip{0pt}
  \setlength\belowdisplayskip{0pt}}

\begin{document}
\title{Layered-Garment Net: Generating Multiple Implicit Garment Layers from a Single Image\thanks{Aggarwal, Wang, Hogue, and Guo are partially supported by National Science Foundation (OAC-2007661).}}
\titlerunning{Layered-Garment Net}
%
\author{Alakh Aggarwal\inst{1} \and
Jikai Wang\inst{1} \and
Steven Hogue\inst{1}\and
Saifeng Ni\inst{2}\and
Madhukar Budagavi\inst{2}\and
Xiaohu Guo\inst{1}}
\authorrunning{A. Aggarwal et al.}
%
\institute{The University of Texas at Dallas, Richardson, TX, US\\
\email{\{alakh.aggarwal,jikai.wang,ditzley,xguo\}@utdallas.edu} \and
Samsung Research America, Plano, TX, US\\
\email{\{saifeng.ni,m.budagavi\}@samsung.com}}
\maketitle              
\begin{abstract}
Recent research works have focused on generating human models and garments from their 2D images. However, state-of-the-art researches focus either on only a single layer of the garment on a human model or on generating multiple garment layers without any guarantee of the intersection-free geometric relationship between them. In reality, people wear multiple layers of garments in their daily life, where an inner layer of garment could be partially covered by an outer one. In this paper, we try to address this multi-layer modeling problem and propose the Layered-Garment Net (LGN) that is capable of generating intersection-free multiple layers of garments defined by implicit function fields over the body surface, given the person's near front-view image. With a special design of garment indication fields (GIF), we can enforce an implicit covering relationship between the signed distance fields (SDF) of different layers to avoid self-intersections among different garment surfaces and the human body. Experiments demonstrate the strength of our proposed LGN framework in generating multi-layer garments as compared to state-of-the-art methods. To the best of our knowledge, LGN is the first research work to generate intersection-free multiple layers of garments on the human body from a single image.

\end{abstract}

\section{Introduction}

\label{sec:intro}

Extracting 3D garments from visual data such as images enables the generation of digital wardrobe datasets for the clothing and fashion industry, and is useful in Virtual Try-On applications. With the limitation on certain classes of garments, it is already possible to generate explicit upper and lower garment meshes from a single image or multi-view images \cite{bhatnagar2019multi,jiang2020bcnet}, to introduce different styles to the garments, such as length, along with varying poses and shapes \cite{patel2020tailornet,su2020deepcloth,corona2021smplicit}, and to transfer the garments from one subject to another~\cite{bhatnagar2019multi}.

However, to the best of our knowledge, none of the existing approaches have the capability of generating multiple \emph{intersection-free} layers of clothing on a base human model where an inner layer of garment could be partially covered by an outer one without any intersection or protrusion. This does not conform to reality because people wear multiple layers of garments in their daily life. The existing techniques either generate a single layer of upper-body cloth (e.g., T-shirt, jacket, etc.) and a single layer of lower-body cloth (e.g., pants, shorts, etc.) without any overlap in their covering regions~\cite{jiang2020bcnet,bhatnagar2019multi}, or generate multiple garment layers, but without any guarantee on their intersection-free geometry~\cite{corona2021smplicit}.

The fundamental challenge here is to ensure intersection-free between multiple garment layers when they overlap. Existing approaches to garment representation are based on \emph{explicit} models, by using either displacement fields over SMPL surface (SMPL+D)~\cite{alldieck2018video,bhatnagar2019multi} or skinned meshes on top of SMPL~\cite{jiang2020bcnet}. However, with explicit mesh representations, it is very difficult to ensure intersection-free between multiple garment layers. SMPLicit~\cite{corona2021smplicit} is an implicit approach that generated multiple layers of garments but does not handle intersections among multiple layers. In this paper, we propose to use a set of \emph{implicit} functions -- signed distance fields (SDF), to represent different layers of garments. The benefit is that the intersection-free condition can be easily enforced by requiring the SDF of the inner layer to be greater than the SDF of the outer one. We call this the \emph{Implicit Covering Relationship} (Sec.~\ref{sec:implicit_covering_relationship}) for modeling multi-layer garments.

There are two challenges associated with the such implicit representation of garments as well as the enforcement of implicit covering relationship: (1) Most of the garments are \emph{open} surfaces with boundaries, while SDF can represent \emph{closed} surfaces only. (2) The implicit covering relationship should only be enforced in those regions where two layers overlap, but how can we define such overlapping regions? In this paper, we solve these two challenges by proposing an implicit function called \emph{Garment Indication Field} (GIF, Sec.~\ref{sec:garment_indication_field}) which successfully identifies those regions where the garment has ``holes'' -- the open regions where the garment does not cover. With such garment indication fields, we not only can enforce the implicit covering relationships between layers but also can extract the open meshes of garments by trimming the closed marching cubes surfaces.

We propose a  \emph{Layered-Garment Net} (LGN), which consists of a parallel SDF subsystem and GIF subsystem, that can take an image of the person as input, and output the corresponding SDF and GIF for each garment layer. Specifically, based on the projection of the query point in image space, we obtain its local image features from the encoded features given by a fully convolutional encoder. Using the local image features and other spatial features of the query point, we train different decoder networks for different layers of garments to predict their SDF and GIF, respectively. The network is trained end-to-end, utilizing a covering inconsistency loss given by GIFs and SDFs of different layers, along with other loss functions to regress the predictions to the ground-truth values. The contributions of this paper can be summarized as follows:
\begin{itemize}
    \item We present a Layered-Garment Net, the first method that can model and generate multiple intersection-free layers of garments and the human body, from a single image.
    \item We enforce an implicit covering relationship among different layers of garments by using multiple signed distance fields to represent different layers, which guarantees that multiple layers of garments are intersection-free on their overlapping regions.
    \item We design garment indication fields that can be used to identify the open regions where the garments do not cover, which can be used to identify the overlapping regions between different layers of garments, as well as to extract open meshes of garments out of the closed surfaces defined by SDF.
\end{itemize}

\section{Related Works}
\label{sec:related}

In this section, we will review the recent works in two areas of research that are related to our work. We consider \textbf{Full Human Body Reconstruction} where the focus is on generating a good quality clothed human model and \textbf{Individual Garment Surface Reconstruction} where the focus is on obtaining individual garments for a human model. 

\paragraph{Full Human Body Reconstruction}
Many recent works generated explicit representations of human body mesh using parametric models for naked human models to handle varying geometry~\cite{loper2015smpl,pavlakos2019expressive,osman2020star}. This allows them to modify the shape and pose of the generated model according to shape parameters $\beta$ and pose parameters $\theta$. The underlying idea is to obtain the parameters $\beta$ and $\theta$ that closely defines the target human body, and apply linear blend skinning using the blend shapes and blending weights to generate the final human body geometry. Bogo et al.~\cite{bogo2016keep} obtained these parameters and fitted a human body model from single unconstrained images. Many deep learning-based methods~\cite{kanazawa2018end,pavlakos2018learning} have since then come up, that estimate the shape and pose parameters of a human model. Smith et al.~\cite{smith2019towards} employed the use of silhouettes from different viewpoints to generate the human body. Subsequently, some research works~\cite{omran2018neural,lassner2017unite} also used semantic segmentation of human parts to ensure more accurate parameter estimation. However, the above-mentioned works only generate naked human models and do not reconstruct clothed human models.

To address this issue, several recent research works have focused on the displacements of a naked body. Alldieck et al.~\cite{alldieck2018video} used frames at some continuous interval from a video of a subject rotating in front of the camera to ensure accurate parameter estimation from different viewpoints and used SMPL+D for clothed human body reconstruction. Such SMPL+D representation uses a displacement vector for the vertices of the naked human body model to represent clothing details and was later used for single image reconstruction~\cite{alldieck2019learning}. Tex2Shape~\cite{alldieck2019tex2shape} was able to obtain better displacement details by predicting the displacement map for a model that aligns with the texture map of the model. Several recent works~\cite{su2020mulaycap} generate explicit dynamic human models. However, since all the above methods are only based on a naked human body model, they cannot generate a human body wearing complex garments like skirts, dresses, long hair, etc. To address these issues, some research works~\cite{varol2018bodynet,zheng2019deephuman} used a volumetric representation of the human body with voxelized grids. Ma et al.~\cite{ma2021scale} obtain the point clouds of clothed humans with varying garment topology. Some recent works~\cite{ma2020learning,wang2020normalgan} also focus on generative approaches for 3D clothed human reconstruction.

There have been some recent works focusing on the implicit clothed body surface representation. Mescheder et al.~\cite{mescheder2019occupancy} used the occupancy field to determine if a point is inside or outside a surface of any object from the ShapeNet~\cite{chang2015shapenet} dataset, and then used a classifier to generate a surface dividing the 3D space into inside/outside occupancy values. They calculated occupancy values for each point of the voxel grid and used marching cubes~\cite{lorensen1987marching} to generate the surface. They do not have to store voxel grid representation or any other mesh information for all the data instances. Different from the occupancy field, Chibane et al.~\cite{chibane2020neural} predicted an unsigned distance field using a neural network, and projected the points back to the surface to generate a point cloud-based surface using the gradient of the distance field at that point, and could be used to further generate a complete mesh surface. Several recent works~\cite{park2019deepsdf,xu2019disn,sitzmann2020metasdf} predicted a Signed Distance Field and used marching cubes~\cite{lorensen1987marching} to generate a mesh surface. This ensures more accurate geometry because of the implicit field's dependence on distance. Based on the above works on implicit fields, PIFu~\cite{saito2019pifu}, PIFuHD~\cite{saito2020pifuhd}, StereoPIFu~\cite{hong2021stereopifu}, GeoPIFu~\cite{he2020geo}, PaMIR~\cite{zheng2021pamir} take a 2D image or depth data of human as input, and after extracting the local encoded image features for a point, they predict the occupancy field of the dressed body. MetaAvatar~\cite{Wang2021NEURIPS} represent cloth-specific neural SDFs for clothed human body reconstruction. Other recent works~\cite{huang2020arch,peng2021neural,peng2021animatable} aim to dynamically handle the reconstruction of animatable clothed human models via implicit representation. Several other works~\cite{liu2021neural,corona2021smplicit,shao2021doublefield,saito2021scanimate} also use implicit fields for 3D human reconstruction. Bhatnagar et al.~\cite{bhatnagar2020combining} combine use base explicitly defined SMPL model to implicitly register scans and point clouds. The method identifies the region between garment and body, however, it does not reconstruct different garment layers. Handling individual garment regions like Garment Indication Field is more complicated. Scanimate~\cite{saito2021scanimate} reconstructs a dynamic human model and utilizes an implicit field for fine-tuning their reconstruction. Instead of supervision, they utilize Implicit Geometric Regularization~\cite{gropp2020implicit} to reconstruct surfaces using implicit SDF in a semi-supervised approach.

\paragraph{Individual Garment Surface Reconstruction}
Instead of simply generating a human body model with displacements, Multi-Garment Net (MGN)~\cite{bhatnagar2019multi} generates an explicit representation of parametric garment models with SMPL+D. Using single or multiple images, it predicts different upper and lower garments that are parameterized for varying shapes and poses. However, MGN cannot produce garments that do not comply with naked human models, like skirts and dresses. TailorNet~\cite{patel2020tailornet} uses the wardrobe dataset from MGN and applies different style transforms like sleeve-length to obtain different styles of garments. DeepCloth~\cite{su2020deepcloth} enable deep-learning based styling of garments. SIZER~\cite{tiwari20sizer} provides a dataset enabling resizing of the garment on the human body. Deep Fashion3D~\cite{zhu2020deep} generates a wardrobe dataset, consisting of complex garment shapes like skirts and dresses. BCNet~\cite{jiang2020bcnet} uses a deformable mesh on top of SMPL to represent garments and proposes a skinning weights generating network for the garments to support garments with different topologies. 
SMPLicit~\cite{corona2021smplicit} obtains shape and style features for each garment layer from the image and uses these parameters to obtain multiple layers of garments, and uses a distance threshold to reconstruct overlapping garment layers. However, they do not guarantee intersection-free reconstruction. GarmentNets~\cite{chi2021garmentnets} reconstructs dynamic garments utilizing Generalized Winding Number~\cite{jacobson2013robust} for occupancy and correct trimming of openings in garment meshes. Their approach, however, does not provide a garment's indicator field.

To the best of our knowledge, none of the existing works can generate overlapping intersection-free multiple layers of garments where an inner layer could be partially covered by an outer one. All existing works on individual garment generation~\cite{bhatnagar2019multi,patel2020tailornet,zhu2020deep,jiang2020bcnet} use explicit mesh representation, making them difficult to ensure intersection-free between different layers. SMPLicit~\cite{corona2021smplicit} does not guarantee intersection-free reconstruction among different layers, especially in overlapping regions. In this paper, we resort to implicit representation and model the multiple layers of garments with signed distance fields (SDF) which makes it easy to enforce the implicit covering relationship among different layers of garment surfaces with the help of a carefully designed implicit garment indication field (GIF). The combination of these two implicit functions, SDF and GIF, makes the modeling and learning of multi-layer intersection-free garments possible.

\input{updated_fig/overview}

\section{The Method}

Given a near-front-facing image of a posed human, we aim to generate the different intersection-free garment surface layers. The reconstructed surfaces should follow a covering relationship between each other and the body. Our proposed Layered-Garment Net (LGN) can generate implicit functions of Signed Distance Field (SDF) and Garment Indication Field (GIF) for different layers of garments over varying shapes and poses. An overview of our approach is given in Fig.~\ref{fig:overview_update}.

\input{fig/cover}
\subsection{Implicit Covering Relationship}
\label{sec:implicit_covering_relationship}
\input{updated_fig/winding}
For two layers of garments $i$ and $j$, let layer $i$ be partially covered by layer $j$. If a point $p$ belongs to their overlapping regions, the SDF values $s_i(p)$ and $s_j(p)$ for the two layers should follow the covering relationship:
\begin{equation}
\label{eq:covering_relationship}
s_j(p) < s_i(p).
\end{equation}
This is illustrated in Fig.~\ref{fig:cover}(left).
The inequality does not hold for all the points in 3D space but only holds for the overlapping region between the two layers. We are only interested in the points near the surface of the garment layer. Let us consider an example where layer $i$ is a pant and layer $j$ is a shirt. The inequality Eq.~\eqref{eq:covering_relationship} should not be satisfied in the leg region of the human body, otherwise, this would result in the generation of a shirt layer on top of the pant layer in the leg region, where the shirt originally does not exist. This problem is shown in Fig.~\ref{fig:cover}(right). Hence, we need an indicator function for both layers, and only ensure that the implicit covering relationship Eq.~\eqref{eq:covering_relationship} holds on points that are related to both layers $i$ and $j$. We call this indicator function for layer $i$ the \emph{Garment Indication Field} (GIF), and denote it as ${h}_i(p)$ (Sec~\ref{sec:garment_indication_field}).

To ensure the network's SDF predictions follow the Implicit Covering Relationship inequality for relevant points $p$, we can define the covering loss for all layers of surfaces for our network as follows:
\begin{equation}
\label{eq:covering_loss}
\mathcal{L}_{cov}(p) = \sum_{j=1}^{N}\sum_{i\in C(j)} h_j(p) * h_i(p) * [max(s_j(p) - s_i(p), 0) + \lambda(s_j(p) - s_i(p))^2],
\end{equation}
where $C(j)$ is the set of layers partially covered by layer $j$. The multiplication with $h_j(p)$ and $h_i(p)$ guarantees that the covering loss only applies to the points in the overlapping region between two layers. The last term regularizes the difference between the two SDFs. We choose $\lambda=0.2$ in all our experiments.

\subsection{Garment Indication Field}
\label{sec:garment_indication_field}

For a garment and a query point $p$, we use its generalized winding number~\cite{jacobson2013robust}, denoted as $W(p)$, to distinguish the open regions from the regions concerned with garment surfaces. Since all the garments are open surfaces, $W(p)$ is equal to $0.5$ at the opening regions. $W(p)>0.5$ for a point inside the surface, and keeps increasing as the point gets farther inside. Similarly, $W(p)<0.5$ for a point outside the surface, and keeps decreasing as it goes further away from the open regions. In far-off regions and outside the surfaces, $W(p)\leq0$. Using different field functions as a function of winding number, we can have different observations as shown in Fig.~\ref{fig:winding}.

\textbf{Observation 1:}\label{sec:winding_occ} $o(p)=W(p)-0.5$ gives the occupancy field for a garment. This has been shown in Fig.~\ref{fig:winding}(b). We call $o(p)$ the \emph{winding occupancy}. This helps us in obtaining the sign of SDF for a non-watertight mesh. Since all garments, in particular, are non-watertight open mesh, for any query point $p$ in 3D space, the distance $d(p)$ to its nearest surface point is essentially an \emph{unsigned} distance because there is no inside/outside for the open surface. Thus we use $o(p)$ to obtain a watertight surface mesh with marching cubes first, then compute a \emph{ground-truth SDF} $s'(p)$ for the watertight garment surface.

\textbf{Observation 2:} $h'(p) = W(p)*(W(p)-w_h)$ gives an indication field of the garment opening region, where $0.5<w_h<1$. As previously discussed, $W(p)$ is greater than $0.5$ inside the opening region and the surface mesh, and it keeps on increasing inside the mesh. Similarly, $W(p)$ is less than $0.5$ outside the opening region and keeps decreasing away from the region outside the mesh. In this paper, we choose $w_h=0.75$ for all garments. For any point that is inside the mesh and away from the garment opening region, $W(p)>0.75$, so $h'(p)$ is positive. Similarly, if it is outside the mesh and away from the garment opening region, $W(p)<0$, so $h'(p)$ is positive too. However, for any point that is located close to the $0.5$-level isosurface, $0<W(p)<0.75$, so $h'(p)$ is negative. In this way, $h'(p)$ indicates the open region of the garment. This can be observed from Fig.~\ref{fig:winding}(c). 

Furthermore, it also follows that $h'(p) - \delta$, for some $\delta \to 0^+$ gives a bound region of the garment closer to the mesh. This has been shown in Fig.~\ref{fig:winding}(d). We observe that, for $\delta=0.01$, we get a good quality bound for this indication field. Thus we define the following function as Garment Indication Field (GIF) for the garment surface:
\begin{equation}
\label{eq:gif}
    \hat{h}(p) = (sign[W(p)*(W(p)-0.75) - \delta]+1)*0.5.    
\end{equation}
Here $\hat{h}(p) = 1$ means the point is in the region close to the garment surface, otherwise $\hat{h}(p) = 0$. Such ground-truth GIF values will be used for enforcing the covering relationship in Eq.~\eqref{eq:covering_loss}.

\subsection{Layered-Garment Net}
\label{sec:layered_garment_net}

Given an input image $I$ of a person, we first obtain the garment segmentation $P$ on the image using Part-Grouping Network~\cite{gong2018instance}. It is possible to obtain different garment masks $g_i$ for garment layer $i$ using the corresponding pixel color. However, the mask $g_i$ may not be complete because of overlap with outer layers. Hence, we train a Garment Mask generator, that takes the incomplete mask $g_i$ and PGN segmentation $P$ as input, and outputs a corrected garment mask $g'_i$.

Like PiFU-HD~\cite{saito2020pifuhd}, we follow a similar pipeline, however, with no requirement for a Fine-Level network, but only a Coarse network for each garment layer. We also use semi-supervised Implicit Geometric Regularization (IGR)~\cite{gropp2020implicit} for fine-tuning SDF prediction on the surface. For layer $i$, we use mask $g'_i$ on input image $I$ and using a Normal Subnetwork masked with $g'_i$, we obtain front and back normals~\cite{saito2020pifuhd}. Let's call the concatenation of masked input image and masked front and back normals for layer $i$ as $N_i$. 

LGN consists of a common SDF Encoder for all layers, that gives feature encoding for $N_i$ as $F_i$. For a given point $p$ in 3D space, we obtain a local pixel feature by orthogonal projection $\pi(p)$ of $p$ on $F_i$, and barycentric interpolation. For the point $p$, we also obtain spatial features like depth. Using the spatial features and local pixel-aligned features, the layer's SDF Decoder $\mathcal{S}_i(.)$ predicts SDF $s_i(p)$.

Similarly, to identify if point $p$ lies in the garment region, we also obtain Indicator Mask $ind_i$ of layer $i$ from PGN $P$ and incomplete mask $g_i$ by training an Indicator Mask generator. Then we train a common GIF Encoder that gives encoding $F'_i$ and the layer's GIF Decoder $\mathcal{H}_i(.)$ to obtain GIF value for layer $i$ as $h_i(p)$.

Mask generators for each garment follow the same architecture as front and back Normal Subnetworks in~\cite{saito2020pifuhd}, i.e. Pix2PixHD network~\cite{wang2018high}. We have a common SDF Encoder and front and back Normal Subnetworks among all garment layers and body layers. A common GIF Encoder is defined for all garment layers. However, we separately define SDF Decoders, GIF Decoders, Garment Mask generators, and Indicator Mask generators for each garment - shirt, pant, coat, skirt, dress.

\begin{equation}
\label{eq:net_pipeline}
    s_i(p) = \mathcal{S}_i(F_i(\pi(p)), \phi(p)), \;
    h_i(p) = \mathcal{H}_i(F'_i(\pi(p)), \phi(p)),
\end{equation}
where the spatial feature $\phi(p)$ here is depth. 

The L1 loss for the generated SDF is formulated as the following $L^1$ norm:
\begin{equation}
\label{eq:sdf_loss}
    \mathcal{L}_{sdf}(p) = \sum_{i=0}^N|s_i(p) - \hat{s_i}(p)|,
\end{equation}
where $\hat{s_i}(p)$ is the ground-truth SDF value for point $p$ from layer $i$, and $N$ is the number of garment layers. 

Similarly, the L1 loss for the predicted GIF is formulated as follows: 
\begin{equation}
\label{eq:loss_gif}
    \mathcal{L}_{gif}(p) = \sum_{i=1}^N|h_i(p)-\hat{h}_i(p)|,
\end{equation}
where $\hat{h}_i(p)$ is the ground-truth GIF for the garment layer $i$ and query point $p$.

We fine-tune network parameters for SDF prediction using Implicit Geometric Regularization (IGR)~\cite{gropp2020implicit}. Loss for IGR is given as follows:

\begin{equation}
\label{eq:igr}
    \begin{split}
    \mathcal{L}_{igr}(p) &= \tau(p)\ell_\mathcal{X}(p) + \lambda(||\nabla_p s_i(p)|| - 1)^2, \\
    \ell_\mathcal{X}(p) &= |s_i(p)| + ||\nabla_p s_i(p) - n_p||, 
    \end{split}
\end{equation}
where $s_i(p)$ is the SDF value at $p$, $\tau(p)$ is an indicator of a point on surface $\mathcal{X}$ and $n_p$ is the surface normal at point $p$.

\subsection{Training and Inference}
\label{sec:training_inference}
We first pre-train the Garment Mask generator and Indicator Mask generators on PGN segmentation and incomplete mask as inputs for each garment category - shirt, pant, coat, skirt, dress.
To train the network, we sample $20,480$ points on the surface of each layer. We add normal perturbation $\mathcal{N}(0,\sigma=5 cm)$ on these points to generate the near-surface samples. We then add random points in 3D space using a ratio of $1:16$ for the randomly sampled points w.r.t. the near-surface samples. These sampled points are used to optimize the SDF prediction of all garment layers and covering loss between each layer of the garment. We similarly sample points from the 0.5 level iso-surface of the ground-truth Garment Indication Fields (GIFs) of each layer garment layer and add normal perturbation $\mathcal{N}(0,\sigma=5 cm)$ on these points to generate garment indications. We add random points in 3D space using a ratio of $1:16$ for the randomly sampled points w.r.t. the garment indicating samples. For GIFs, we add additional points along the edges of ground truth mesh to obtain accurate trimming.

Given an input image $I$, we first obtain PGN segmentation image $P$ which contains different garments in the image. For each garment, we obtain their incomplete masks $g_i$. Using $P$ and $g_i$ for each layer, we obtain garment masks $g'_i$ and $ind_i$. We leave out the indicator mask prediction for the body layer since it is not required to obtain GIF for the body. It is assumed that the GIF value for the body layer is $1$ at any point. For all the near-surface sampled points, we calculate the ground-truth SDF values $\hat{s}_i$ for each layer $i$ as explained in~\ref{sec:winding_occ}. The encoder and decoder are warmed-up by training with the loss $\mathcal{L}_{sdf}$ and $\mathcal{L}_{igr}$ as defined in Eq.~\eqref{eq:sdf_loss} and Eq.~\eqref{eq:igr}. We also calculate ground-truth GIF values $\hat{h}_i$ for each layer $i>0$ with the loss $\mathcal{L}_{gif}$ as defined in Eq.~\eqref{eq:gif}. Using the predicted $h_i$ values and the predicted $s_i$ values for all the sampled points, the network is fine-tuned with the covering loss as defined in Eq.~\eqref{eq:covering_loss}. This ensures that the output SDF values follow the covering relationship inequality as defined in Eq.~\eqref{eq:covering_relationship}. For all the garments indicating sampled points, we calculate their ground-truth GIF values for each layer of the garment. Using the ground-truth GIF values for each layer, the GIF value prediction is optimized.

During the inference, after obtaining the SDF values for each layer, we use marching cubes to obtain its triangle mesh. Then, we apply a trivial post-processing step using predictions, to update SDF values. For a given point $p$ where GIF of both layers $i$ and $j$ overlap:
if $s_j > s_i - \epsilon\text{, }s_j = s_i - \epsilon$
where $\epsilon$ is a very small number. Experimentally, we use $\epsilon$ to be $1e-3$. Finally, all the triangular meshes obtained for each layer are trimmed by the predicted GIF values on the vertices of the mesh. To trim the garment opening regions, the triangles which have different signs of GIF values for its three vertices are selected, and the triangle is trimmed by linearly interpolating GIF values over each edge. Thus, we finally obtain multiple layers of garments along with the reconstructed Layer-$0$ body that follows the covering relationship.

For both training and inference, we rely on covering the relationship manually specified with the input image. Different garment layers are then obtained from the output of LGN by satisfying the covering relationship.

\section{Experiments}
\subsection{Dataset and Implementation Details}
\textbf{Dataset Preparation.} Our multi-layer garment dataset is constructed from 140 purchased rigged human models from AXYZ~\cite{axyz}. For each rigged model, we first perform SMPL~\cite{loper2015smpl} fitting to obtain its body shape and pose parameters. We generate eight images from different views for each human model and run semantic segmentation on each image with Part Grouping Network (PGN)~\cite{gong2018instance}. Using the fitted SMPL, we obtain those segmentations on the SMPL surface and map them to the UV texture space of SMPL. This enables us to perform texture stitching~\cite{alldieck2018detailed} to generate the segmentation texture map. By projecting the texture segmentation onto the 3D human model, we obtain the segmentation of different 3D garment meshes, followed by minor manual corrections on some garment boundaries. Our processed garments include the categories of Shirt, Coat, Dress, Pant (long and short), and Skirt, while Shirt/Coat/Dress all contain three subcategories of no-sleeve, short-sleeve, and long-sleeve. Detailed statistics of the processed garments are provided in the supplementary document.

\input{tab/results} 
Using different garments, we synthesize around $12,000$ different combinations of multi-layer garments on top of a layer-0 SMPL body, in $7$ different poses. When combining different garment types, we follow the assumption that the length of sleeves for the inner layer should NOT be shorter than that of the outer layer. Otherwise, the sleeves of the inner layer are covered by the outer garment and there is no visual clue to tell its length. We then use this combination of generated garment models with a layer-0 body to train our LGN. We use the synthesized combinations of multi-layer garments as the training set. The geometries of garments are corrected to make sure no intersections exist and the different layers of garment follow the covering relationship. For testing, we use BUFF~\cite{Zhang_2017_CVPR}  and Digital Wardrobe~\cite{bhatnagar2019multi} datasets. The dataset preparation details are discussed in the supplementary document.

\textbf{Implementation Details.} The base architecture of our LGN is similar to that of PIFu~\cite{saito2019pifu} and PIFu-HD~\cite{saito2020pifuhd} since they also predict implicit fields aligned with image features. For SDF Subsystem, we first obtain garment masks from PGN segmentation of input image using Garment Mask Generators and obtain masked front and back normals using Normal Subnetworks, which are Pix2Pix-HD~\cite{wang2018high} networks. We use 4 stacks of Stacked Hourglass Network (HGN)\cite{newell2016stacked} to encode the image features from the concatenation of normals and image. From spatial features from points and local encoded features by performing bi-linear interpolation of projected points on image feature space, different Multi-layer Perceptron (MLP) decoder layers predict SDF values for each layer of the garment, with layer 0 being the human body. Similarly, for GIF Subsystem, Indicator Mask Generators obtain GIF masks. 4 stacked-HGN encodes image features from the concatenation of PGN segmentation and indicator mask. Then, GIF is predicted using GIF Decoders. To optimize the network, we first pre-train Mask Generators. Then we individually train SDF Subsystem and GIF Subsystems. Thereafter, we use the covering loss to fine-tune SDF prediction to avoid the intersection, and GIF to ensure appropriate trimming of the open region on garment surfaces, and a consistent multi-layer covering relationship. We evaluate our methods and test with various state-of-the-art approaches on mainly two areas - 3D Clothed Human Reconstruction and Individual Garment Reconstruction. The quantitative comparison of 3D Clothed Human Reconstruction of our method with BCNet~\cite{jiang2020bcnet}, PiFU-HD~\cite{saito2020pifuhd} and SMPLicit~\cite{corona2021smplicit} are shown in the supplementary document. We omit comparison with MGN~\cite{bhatnagar2019multi}, Octopus~\cite{alldieck2018detailed} and PiFU~\cite{saito2019pifu} because of the availability of better reconstruction methods.

\subsection{Quantitative Comparisons}
We compare our methods with the state-of-the-art (Tab.~\ref{tab:res}) approaches on three publicly available datasets: Digital Wardrobe Dataset~\cite{bhatnagar2019multi}, SIZER Dataset~\cite{tiwari20sizer} and BUFF Dataset~\cite{Zhang_2017_CVPR}. We use Digital Wardrobe Dataset and SIZER Dataset to compare individual garment reconstructions, and BUFF Dataset~\cite{Zhang_2017_CVPR} to compare full human body reconstruction. It is to be noted that since the datasets mentioned consist of only 2 layers of garments -- upper and lower, we cannot make a comparison with them on multi-layer garment reconstruction. Please also note, we do not use BCNet~\cite{jiang2020bcnet} data set to have a fair comparison with BCNet. Also, we are unable to compare our results with DeepFashion3D~\cite{zhu2020deep} data set because the dataset only consists of garments and no human body.

\begin{minipage}[h]{.9\linewidth}
    \begin{minipage}{.55\linewidth}
        \begin{figure}[H]
            \centering
            \begin{tabular}{|c|c|c|c|c|c|}
            \hline
                \textbf{Method} & \textbf{Im1} & \textbf{Im2}  & \textbf{Im3}  & \textbf{Im4}  & \textbf{Im5}  \\\hline
                SMPLicit & 12.9 & 32.7 & 1.67 & 21.3 & 18.3
                \\
                Ours & 0.34 & 0 & 0 & 0.16 & 0\\
                \hline
            \end{tabular}
            \includegraphics[width=0.9\textwidth]{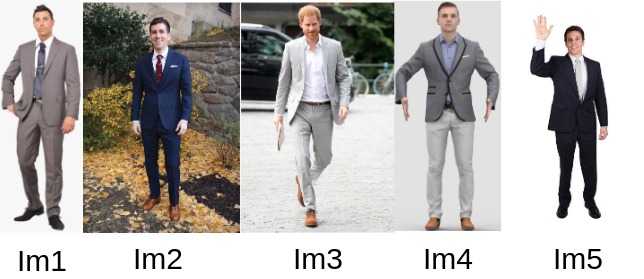}
            \caption{Penetration depths (in cm).}
            \label{fig:max_pen}
        \end{figure}
    \end{minipage}
    \quad
    \begin{minipage}{.5\linewidth}
        \begin{figure}[H]
            \begin{center}
                \includegraphics[width=0.9\linewidth]{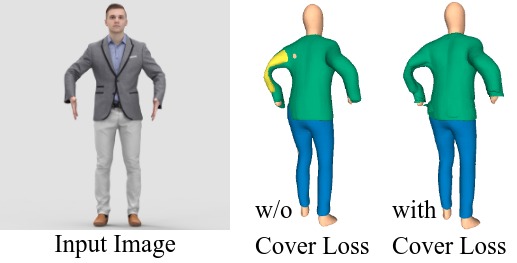}
            \end{center}
            \caption{Cover Loss Finetuning
            }
            \label{fig:coverlossfinetune}
        \end{figure} 
    \end{minipage}
\end{minipage}

\subsubsection{Individual Garment Reconstruction}
To compare our method with the state-of-the-art garment reconstruction approaches~\cite{corona2021smplicit,jiang2020bcnet}, we select $96$ models from Digital Wardrobe Dataset~\cite{bhatnagar2019multi} and $97$ models from SIZER Dataset~\cite{tiwari20sizer}. We use segmented Upper and Lower garments available with Dataset for comparison. We calculate the Mean P2S Error per garment between reconstructed garments and their ground-truth counterpart and observe the performance of our approach with other approaches in Tab.~\ref{tab:res}~A(i)\&(ii). Our model outperforms state-of-the-arts on Digital Wardrobe Dataset~\cite{bhatnagar2019multi}. For SIZER~\cite{tiwari20sizer}, BCNet performs better due to assuming reconstruction of $2$ layer garments only, since segmentation for $3$ layer of garments does not exist in the data set.

In Fig.~\ref{fig:max_pen}, we calculate the Maximum Penetration Depth between different reconstructed garment layers and make a comparison with SMPLicit~\cite{corona2021smplicit}. It can be seen that our work outperforms the state-of-the-art in this case.

\subsubsection{Full Human Body Reconstruction}
\input{fig/res/human_qual}
\input{fig/res/cover_out}
We show in Tab.~\ref{tab:res}~B the comparison of our method with the state-of-the-arts on full human body reconstruction, on $26$ models consisting of different subjects and clothes from BUFF Dataset~\cite{Zhang_2017_CVPR}. We calculate Chamfer distance and Point-to-surface (P2S) error between ground-truth human models and reconstructed full body surface. We do not compare with SMPLicit, because they have no method for full body reconstruction. Please note that we encounter lower results in this case than state-of-the-arts because we do not focus on accurate naked body (layer 0) reconstruction.

\subsection{Qualitative Results}
We compare the reconstruction quality of garment surfaces on the human body in Fig.~\ref{fig:human_qual}. We can observe that our method (LGN) reconstructs a more detailed 3D human body than state-of-the-art explicit model reconstruction methods like BCNet, showing the effectiveness of implicit model reconstruction in comparison explicit approach. Also, SMPLicit generates a very coarse structure and loses many finer details for the clothes on the human body. Since we can generate individual implicit garment surfaces, we can retain finer details, especially between different layers. Since our networks are fine-tuned with IGR Loss, we reconstruct garments of similar quality to PIFu-HD without using fine-level networks.

In Fig.~\ref{fig:cover_out}, we further show different challenges faced in the reconstruction of different garment surfaces. In the top row, we show the effect of covering relationship on multiple layers of garment reconstruction, specifically for the Shirt and Pant layers. From the given image, we expect the Pant layer to cover the Shirt layer without intersections. However, BCNet generates Shirt covering Pants, according to their pre-defined template. On the other hand, SMPLicit completely misses covering relation. In the bottom row, we show the reconstruction of the Coat layer above the Shirt layer as in the image. We expect two layers of garment reconstruction for the upper body. However, since BCNet is based on an explicit reconstruction of garments based on a displacement map on the SMPL body, it cannot reconstruct two-layer geometry for the upper body. Since SMPLicit does not guarantee intersection-free reconstruction, we find intersections between Shirt and Coat layer. Since the results of our LGN satisfy the covering relationship in Eq.~\eqref{eq:covering_relationship}, we get the expected output of garments in both cases.

In Fig.~\ref{fig:coverlossfinetune}, we show how Covering Loss affects the reconstruction output. Without covering loss finetuning, inner layers intersect with outer layers.

\section{Conclusion, Limitation, and Future Work}

We introduce a novel deep learning-based approach that reconstructs multiple non-intersecting layers of garment surfaces from an image. Our approach enforces the implicit covering relationship between different garment layers and the human body and identifies overlapping regions of different garment layers, as well as extract open (non-watertight) meshes. To the best of our knowledge, Layered-Garment Net (LGN) is the first approach that can handle the intersection-free reconstruction of multiple layers of garments from a single image.

Our approach currently does not handle color information, since obtaining good texture for multiple reconstructed layers is difficult. Other neural implicit functions (e.g. Neural Radiance Fields) can address this issue. Our approach does not handle more challenging geometries consisting of manifold garment surfaces and details like pockets, hoodies, collars, etc., and some challenging poses, like limbs close to the body, etc. Also, since the naked human body model was not the focus of this work, the current approach does not handle the detailed full-body reconstruction. These issues can be a major improvement for future work.

\newpage
%
%
\bibliographystyle{splncs}
\bibliography{samplepaper}
\end{document}

%% file: updated_fig/overview.tex
\begin{figure}[t]
\begin{center}
\includegraphics[width=0.9\linewidth]{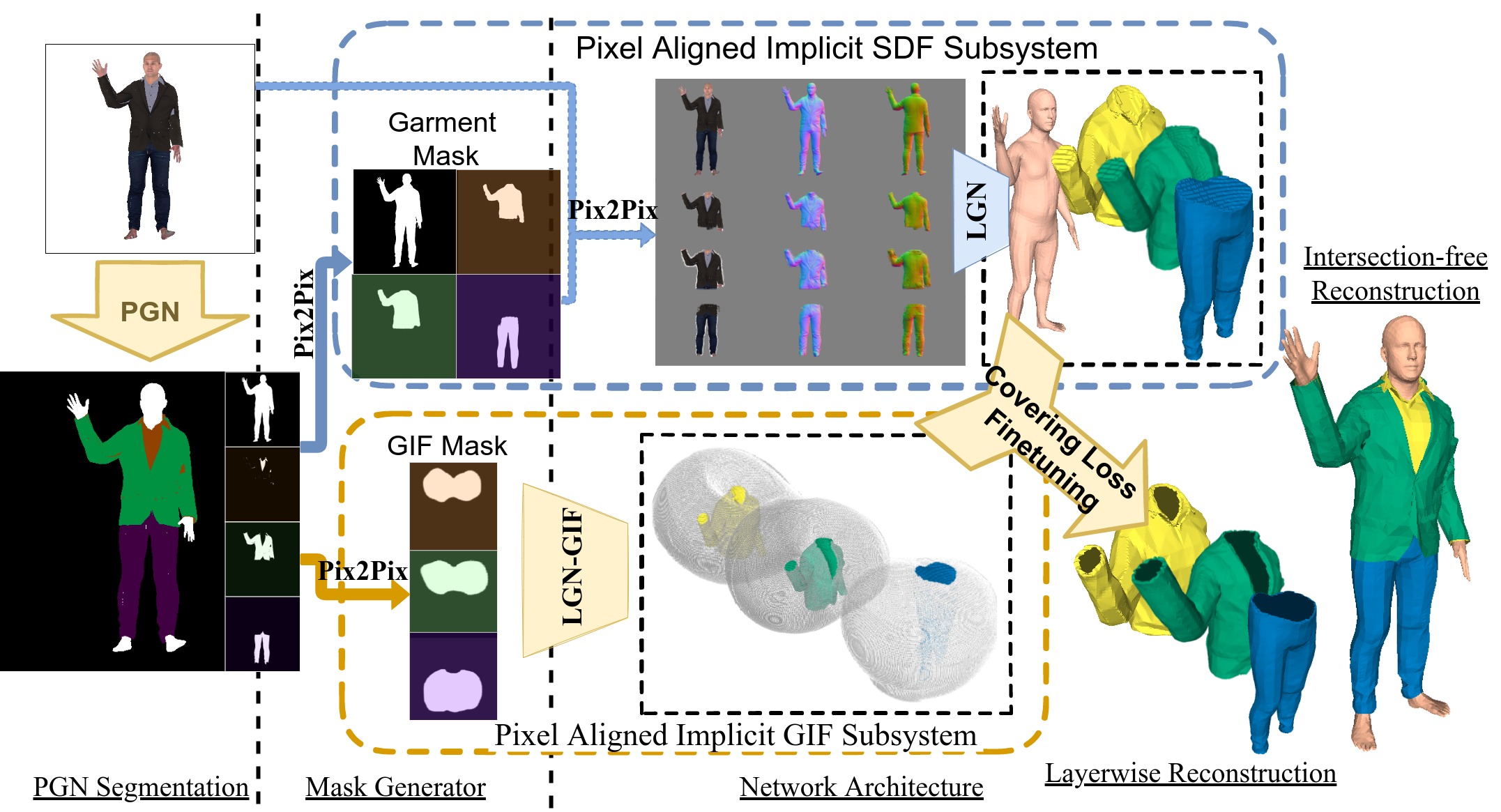}
\end{center}
\caption{
%
Given an input image, our method first obtains PGN~\cite{gong2018instance} segmentation and incomplete masks for each garment from segmentation. Then the complete masks of garments are generated by Pix2Pix-HD Garment Mask generator. Similarly the indicator masks for garments are generated by Pix2Pix-HD Indicator Mask generator. Using the masked input image, the front and back normals are obtained using Pix2Pix-HD Normal Subnetwork. Then an Encoder and Decoders of LGN network use masked images and normals and predicts SDF value $s_i(p)$ of layer $i$ for any point $p$ in 3D space. LGN-GIF further uses Indicator Masks $ind_i$ to obtain GIF value $h_i(p)$ of layer $i$ for any point $p$ in 3D space. Finally, LGN is fine-tuned with the covering loss in Eq.~\eqref{eq:covering_loss} to avoid intersection among different layers.
}
\label{fig:overview_update}
\end{figure}

%% file: fig/cover.tex
\begin{figure}[h]
\begin{center}
\includegraphics[width=0.45\textwidth]{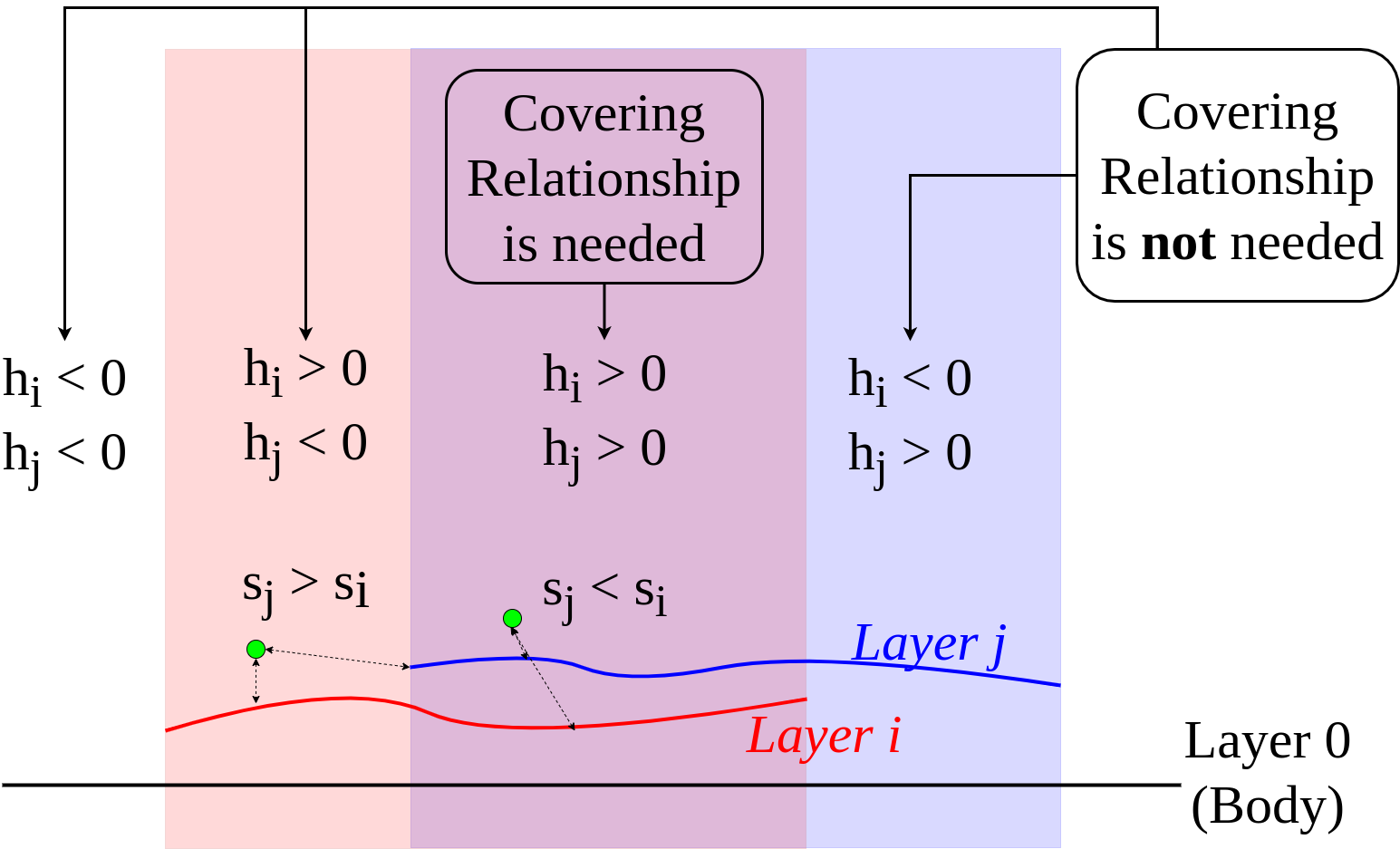}
\includegraphics[width=0.45\textwidth]{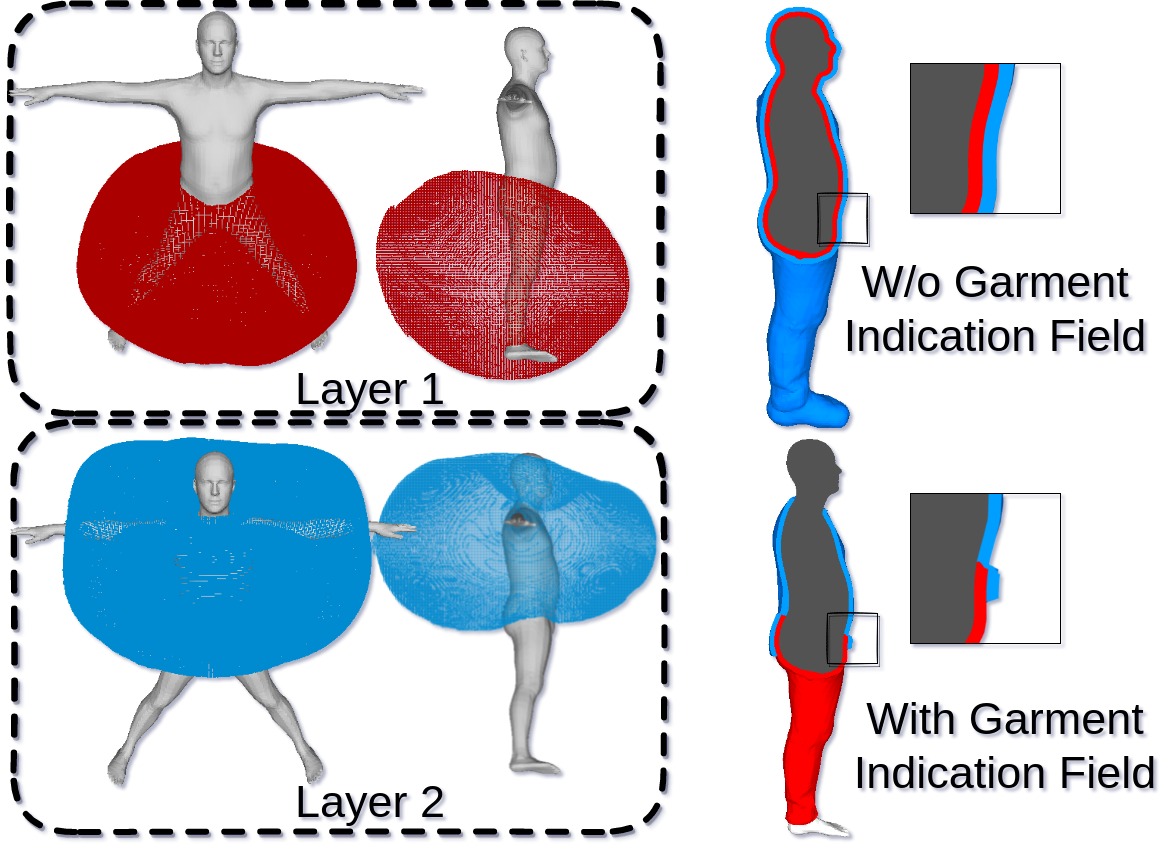}
\end{center}
\caption{
(Left) For a point $p$ associated with  two layers of surfaces $i$ and $j$, where layer $i$ is partially covered by layer $j$, it should satisfy $s_i(p) > s_j(p)$ in their overlapping region where $h_i(p)>0$ and $h_j(p)>0$. For other regions, this relationship may not satisfy. (Right) Garment Indication Fields (GIF) of an inner layer pant and an outer layer shirt are used to constrain the covering relationship only in their overlapping region. Without GIF, the outer layer would completely cover the inner layer since $s_i(p) > s_j(p)$ would be enforced everywhere.
}
\label{fig:cover}
\end{figure}

%% file: updated_fig/winding.tex
\begin{figure}[t]
\begin{center}
\includegraphics[width=0.8\linewidth]{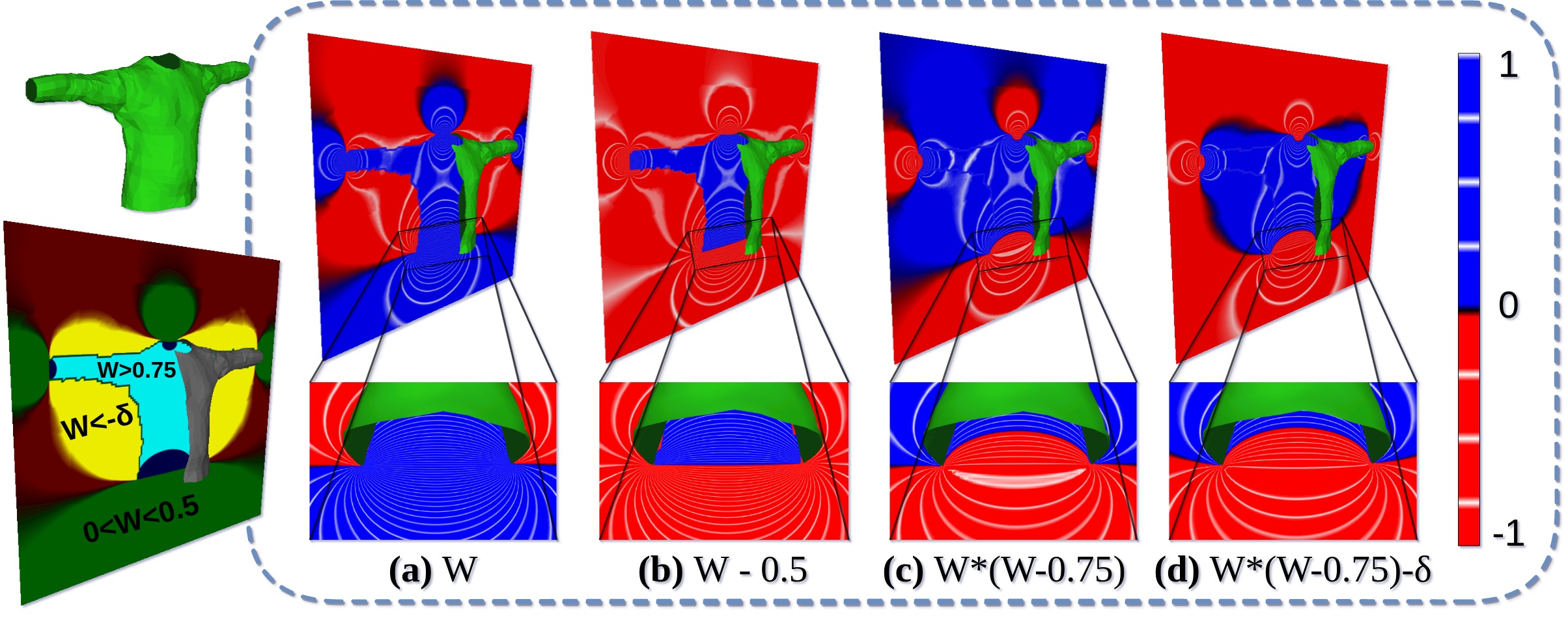}
\end{center}
\caption{
For a given garment mesh (non-watertight), we show the cross-section views of the following fields: (a) Generalized Winding Number $W$, (b) Winding Occupancy $W-0.5$, (d) ``Hole'' Region Indication $W*(W-0.75)$, and (e) Garment Indication Field $W*(W-0.75)-\delta$. Transition from (c) to (d) allows concise bound for GIF, which will not intersect with nearby body surfaces.}
\label{fig:winding}
\end{figure}

%% file: tab/results.tex
\vspace{-4mm}
\begin{table}[htbp]
    \centering
    A(i)
    \begin{tabular}{|c|c|}
    \toprule
        \textbf{Model} & \textit{P2S} \\
        \midrule
        BCNet & 9.75\\
        SMPLicit & 9.12\\
        Ours & \textbf{\underline{9.09}}\\
        \bottomrule
    \end{tabular}
    A(ii)
    \begin{tabular}{|c|c|}
    \toprule
        \textbf{Model} & \textit{P2S} \\
        \midrule
        BCNet & \textbf{3.84}\\
        SMPLicit & 6.01\\
        Ours & \underline{4.04}\\
        \bottomrule
    \end{tabular}
    \quad
    B
    \begin{tabular}{|c|c|c|}
    \toprule
        \textbf{Model} & \textit{Chamfer} & \textit{P2S} \\
        \midrule
        PiFU-HD & \textbf{1.22} & \textbf{1.19}\\
        BCNet & 1.93 & 1.96\\
        Ours & \underline{2.75} & \underline{2.6}\\
        \bottomrule
    \end{tabular}
    \caption{Comparison results (in $cm$) for \textbf{A} per-garment Point-to-Surface on (i)Digital Wardrobe~\cite{bhatnagar2019multi}, (ii) SIZER~\cite{tiwari20sizer}, and \textbf{B} Full body reconstruction on BUFF Dataset~\cite{Zhang_2017_CVPR}}
    \label{tab:res}
\end{table}
\vspace{-4mm}

%% file: fig/res/human_qual.tex
\begin{figure}[h]
  \centering
  \includegraphics[width=0.8\linewidth]{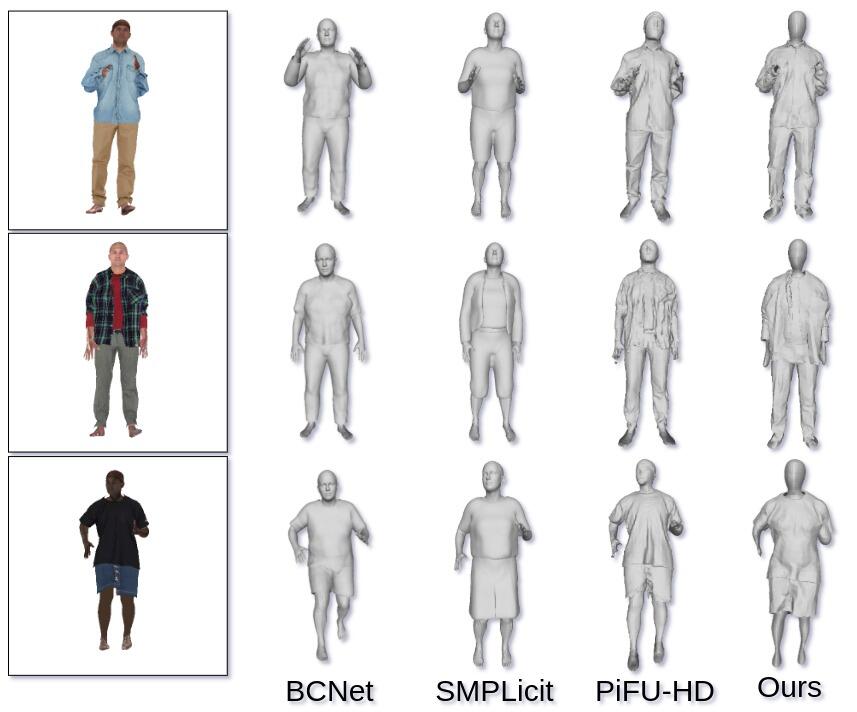}

   \caption{From left to right, qualitative comparison of full-body reconstruction on 3D clothed human from ground truth (left), and results from BCNet~\cite{jiang2020bcnet}, SMPLicit~\cite{corona2021smplicit}, PIFu-HD~\cite{saito2020pifuhd} and Ours (LGN).}
   \label{fig:human_qual}
\end{figure}

%% file: fig/res/cover_out.tex
\begin{figure}[h]
  \centering
  \includegraphics[width=0.7\linewidth]{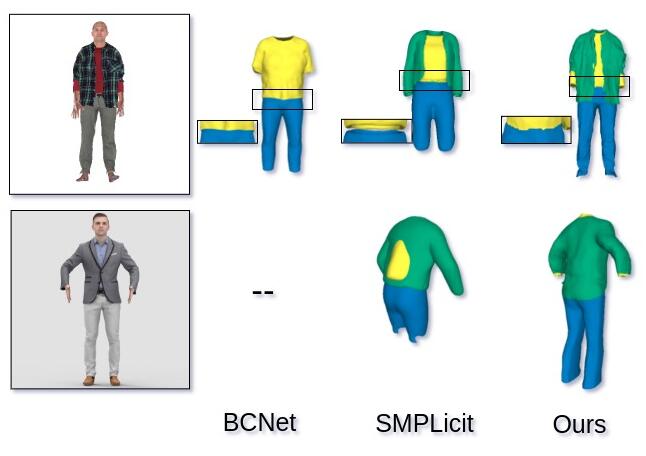}

   \caption{Comparison between the reconstruction results of ours, BCNet~\cite{jiang2020bcnet} and SMPLicit~\cite{corona2021smplicit}. Our reconstruction achieves intersection-free between different layers by satisfying the implicit covering relationship, while BCNet cannot reconstruct multi-layered garment structure, and the result from SMPLicit does not have such guarantee and has clear intersections between different layers. 
   }
   \label{fig:cover_out}
\end{figure}